\documentclass[journal,transmag]{IEEEtran}

\usepackage{times}
\usepackage{latexsym}
\usepackage{graphicx}              
\usepackage{multirow}
\usepackage{bm}
\usepackage{times}
\usepackage{amsmath}
\usepackage{enumerate}
\usepackage{CJK}
\usepackage{url}
\usepackage{booktabs} 
\usepackage{footnote}
\usepackage{tabularx}
\usepackage{algorithm}
\usepackage{algorithmic}
\usepackage{float}
\usepackage{color}
\usepackage{arydshln}
\usepackage{bm}
\usepackage{amssymb}
\usepackage{colortbl} 
\usepackage{multirow} 
\usepackage{multicol}
\usepackage{colortbl}  
\usepackage{cite}
\usepackage{caption} 
\ifCLASSINFOpdf
\else
\fi
\begin{document}
\begin{CJK*}{UTF8}{gbsn}

\title{Conversational Semantic Role Labeling}


\author{\IEEEauthorblockN{
Kun Xu\IEEEauthorrefmark{1},
Han Wu\IEEEauthorrefmark{2},
Linfeng Song\IEEEauthorrefmark{1},
Haisong Zhang\IEEEauthorrefmark{1},
Linqi Song\IEEEauthorrefmark{2},
and Dong Yu\IEEEauthorrefmark{1}}
\IEEEauthorblockA{\IEEEauthorrefmark{1}Tencent AI Lab}
\IEEEauthorblockA{\IEEEauthorrefmark{2}City University of Hong Kong}}
% <-this % stops an unwanted space
% \thanks{Manuscript received December 1, 2012; revised August 26, 2015. 
% Corresponding author: M. Shell (email: http://www.michaelshell.org/contact.html).}}

% \markboth{Journal of \LaTeX\ Class Files,~Vol.~14, No.~8, August~2015}%
% {Shell \MakeLowercase{\textit{et al.}}: Bare Demo of IEEEtran.cls for IEEE Transactions on Magnetics Journals}



% for Transactions on Magnetics papers, we must declare the abstract and
% index terms PRIOR to the title within the \IEEEtitleabstractindextext
% IEEEtran command as these need to go into the title area created by
\IEEEtitleabstractindextext{%
\begin{abstract}
Semantic role labeling (SRL) aims to extract the arguments for each predicate in an input sentence.
Traditional SRL can fail to analyze dialogues because it only works on every single sentence, while ellipsis and anaphora frequently occur in dialogues.
To address this problem, we propose the conversational SRL task, 
where an argument can be the dialogue participants, a phrase in the dialogue history or the current sentence.
As the existing SRL datasets are in the sentence level, we manually annotate semantic roles for 3,000 chit-chat dialogues (27,198 sentences) to boost the research in this direction.
Experiments show that while traditional SRL systems (even with the help of coreference resolution or rewriting) perform poorly for analyzing dialogues, modeling dialogue histories and participants greatly helps the performance, indicating that adapting SRL to conversations is very promising for universal dialogue understanding.
Our initial study by applying CSRL to two mainstream conversational tasks, dialogue response generation and dialogue context rewriting, also confirms the usefulness of CSRL.
\end{abstract}}

% \begin{IEEEkeywords}
% IEEE, IEEEtran, IEEE Transactions on Magnetics, journal, \LaTeX, magnetics, paper, template.
% \end{IEEEkeywords}}

\maketitle

\IEEEdisplaynontitleabstractindextext

% \fi
\IEEEpeerreviewmaketitle

\section{Introduction}
% Semantic role labeling (SRL) \cite{palmer2010semantic} aims to identify the predicate-argument (PA) structures of an input sentence, where the PA structures are supposed to capture the main semantic information of \emph{who} did \emph{what} to \emph{whom}. 
% % Figure \ref{fig:example}(a) shows an example.\\
% Formally, this task involves two sub-tasks: (1) detecting the predicates
% % (e.g., ``{\small 指导} (directed)'')
% ; (2) identifying and assigning their corresponding arguments to semantic roles. 
% % (e.g., ``{\small 克里斯·巴克、珍妮弗·李} (Chris Buck and Jennifer Lee)'' and ``{\small 冰雪奇缘2} (Frozen II)'' are \textbf{A0} role and \textbf{A1} role of the predicate ``{\small 执导} (directed)'').
% Since being proposed, SRL has become a standard module in many major NLP toolboxes \cite{manning2014stanford,2018_lrec_cogcompnlp,gardner2018allennlp}.
% In addition to that, SRL has also been shown crucial for many downstream NLP applications, such as machine translation \cite{marcheggiani2018exploiting}, question answering \cite{khashabi2018question,fitzgerald2018large,dua2019drop,mihaylov2019discourse,zhang2019explicit} and summarization \cite{yan2014srrank}.

Recent years have witnessed increasing attentions of conversation-based tasks, such as dialogue response generation \cite{li2017dailydialog,zhang2018personalizing,dinan2018wizard,wu-etal-2019-proactive,zhou-etal-2020-kdconv}, task-oriented dialogue modeling \cite{henderson-etal-2014-second,mrkvsic2017neural,budzianowski2018multiwoz} and conversational question answering \cite{choi2018quac,reddy2019coqa}.
As a central problem, ellipsis and anaphora frequently occur in human conversations, creating additional challenges for dialogue understanding.
Specifically, ellipsis corresponds to the situations when a phrase or a clause that has been mentioned in the previous context is omitted for simplicity, and anaphora happens when a mention is replaced by a pronoun to avoid repetition.
In addition to English, previous work \cite{kim2000subject} has shown that both phenomena can be much severer for pro-drop languages, including Japanese, Chinese and Korean.

\begin{figure}[t!]
    \centering
    \includegraphics[width=1.0\linewidth]{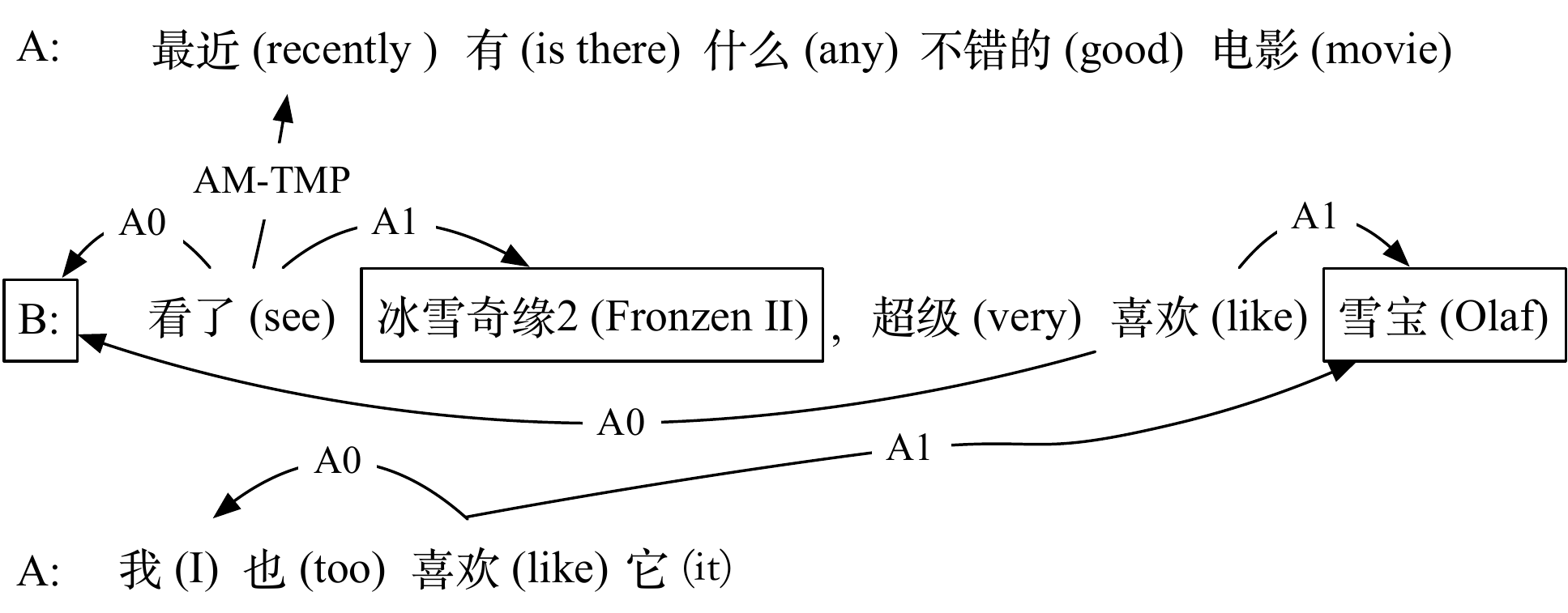}
    \caption{A conversational SRL example. It covers all the core relations, while traditional SRL can only cover the relations indicated by black arrows. One slashed line indicates a coreference relation.}
    \label{fig:example}
\end{figure}

One potential solution is to utilize SRL, a standard tool for semantic analysis inside most traditional NLU toolboxes \cite{manning-EtAl:2014:P14-5,gardner2018allennlp}, to extract valuable information for the end task.
Figure~\ref{fig:example} shows a $3$-turn dialogue example in Chinese with SRL annotations, where relations like ``{\small 雪宝} (Olaf)'' being the ``A1'' argument of the predicate ``{\small 喜欢} (like)'' in the second turn clearly indicate the core subject-predicate-object relations (also fine-grained sentiment expressed by ``{\small 喜欢} (like)'').
However, traditional SRL can only work on every single sentence, thereby they may fail to capture the critical information across turns (utterances) for these conversational tasks.
For instance, several semantic roles, such as ``A0'' and ``AM-TMP'' for the predicate ``{\small 看了} (see)'' in the second sentence, can be missed by traditional SRL, because these arguments are dropped from the sentence.
Besides, it can be challenging to predict the relation between ``{\small 雪宝} (Olaf)'' with the predicate ``{\small 喜欢} (like)'' in the third turn,
because that involves combining SRL (``{\small 它} (it)'' being ``A1'' of ``{\small 喜欢} (like)'') with coreference resolution (``{\small 它} (it)'' referring ``{\small 雪宝} (Olaf)'').

To address this, previous approaches \cite{elgohary-etal-2019-unpack,su-etal-2019-improving} first perform sentence rewriting before conducting standard SRL, where sentence rewriting recovers dropped components and pronouns of coreference by \emph{copying} from dialogue histories.
For example, in Figure \ref{fig:example}, the second turn can be rewritten as ``{\small \underline{我(I) 最近(recently)} 看了 (see) 冰雪奇缘2 (Frozen II)，\underline{我(I)} 超级 (very) 喜欢 (like) 雪宝 (Olaf)}'', where recovered components are with underlines.
Based on the rewritten sentence, missing cross-sentence information is supposed to be captured by standard SRL.
However, the performance of the state-of-the-art rewriting models are still far from satisfactory for out-of-domain scenarios.
%and their rewriting outputs are likely to repeat or miss some words, which may even result in nonsensical sentences.
This is probably because sentence rewriting is quite ``subjective'', as annotators usually have different ideas on the ideal rewriting outputs, resulting in high variance for the annotations.

In this paper, we introduce a new task, namely \textit{conversational semantic role labeling} (CSRL), to directly model the predicate-argument structures across the entire conversation instead of an individual sentence.
%Modeling in this way, the error propagation suffered by previous solutions can be avoided.
Our motivation behind proposing the task comes from one observation: while ellipsis and coreference frequently happen in each dialogue utterance, most dropped or referred components have been mentioned in the dialogue history, and very little information is missing from the whole dialogue.
Following this observation, when analyzing a predicate in the latest dialogue turn, a CSRL model needs to consider not only the current turn but also its previous history sentences to search for potential arguments.
Taking Figure \ref{fig:example} as an example, a system has to jointly consider the first two turns when analyzing the second turn (by \emph{speaker B}), in order to capture all semantic roles.
Compared with standard SRL that needs sentence rewriting or coreference resolution as a preprocessing step for analyzing dialogues, CSRL directly processes dialogues and can avoid error propagation.

There has been no publicly available data that contains the predicate-argument (PA) annotations beyond the sentence level.
Therefore, as the first step to conversational SRL, we manually annotated DuConv \cite{wu-etal-2019-proactive}, a popular dataset of Chinese chit-chat dialogues, with conversation-level semantic roles. 
% As a pro-drop language, both ellipsis and anaphora frequently occur in Chinese, making it a great test-bed for studying conversational SRL.
Our corpus contains $3,000$ dialogue sessions, including $27,198$ utterances, $33,673$ predicates and their arguments.
Besides, we respectively annotate $200$ and $300$ dialogue sessions from the corpus by \cite{zheng2019personalized} and an in-house dataset as out-of-domain test sets to verify model robustness.

Experimental results demonstrate that our CSRL parser significantly outperforms the baselines trained on the standard CoNLL-2012 benchmark \cite{pradhan2013towards}, though the standard benchmark contains four times more training instances than ours (roughly $117,089$ vs. $27,198$).

In addition, we make initial attempts that apply our CSRL parser on two dialogue downstream tasks, i.e., dialogue rewriting and dialogue generation.
In particular, we first utilize our CSRL parser trained on our annotated dataset to extract PA structures, before incorporating this information into rewriting and generation baselines.
Experiments on several benchmarks show that CSRL can significantly improve the state-of-the-art performances for both tasks.

In summary, our contributions are:
\begin{enumerate}
    \item We propose a new task (i.e., CSRL) for modeling the PA structures in multi-turn dialogues.
            For this purpose, we collect an annotated dataset of high quality and two out-of-domain datasets for robust testing, all of which can benefit future related research\footnote{\url{https://github.com/syxu828/CSRL_dataset}.}.
    \item We conduct detailed analysis to show that our model trained with our CSRL data largely outperforms the baselines trained on standard SRL data, and the improvements are consistent across different domains.
    %propose a simple yet effective transformer-based model outperforming existing state-of-the-art          standard-SRL baselines.
            Experimental results on out-of-domain datasets also demonstrate the robustness of our model for domain transfer.
    \item Experimental results on the rewriting and dialogue generation datasets show that our CSRL parser could significantly improve existing baseline models by incorporating the predicate-argument information.
\end{enumerate}

\section{Related Work}
\textbf{Semantic role labeling beyond sentence level.}
Until now, SRL is typically considered as a sentence-internal task, and the major SRL benchmarks \cite{carreras2005introduction,pradhan2013towards} only contain sentence-level annotations.
A partial reason can be that previous manual annotation projects choose to label isolated sentences to avoid wasting expensive human resources on rare long-range cases \cite{ruppenhofer2009semeval}.
A few later efforts \cite{ruppenhofer2009semeval,gerber2012semantic,roth2015inducing} propose \emph{discourse-based implicit argument linking} as a remedy for long-range situations.
Specifically, they introduce an \emph{implicit argument} ``\emph{null}'' in the target sentence for a cross-sentence argument, linking ``\emph{null}'' to the argument.
Conversely, we directly annotate the cross-sentence arguments for each predicate.
These efforts only provide tens or hundreds of labeled instances while we annotated more than 27K sentences, enabling the training and evaluation for state-of-the-art neural models.

\textbf{Dialogue understanding.}
It is essential to extract user intent for building successful dialogue systems.
For task-oriented scenarios \cite{wen2017network}, the goal of dialogue understanding is definite, i.e., identifying the user intent (e.g. \emph{booking flight}) and its corresponding semantic slots (e.g. \emph{from\_city} and \emph{to\_city}).
In contrast, people lack consensus on what information can be helpful for modeling chit-chat dialogues, adding extra difficulties for building intelligent chatbots.
Most previous efforts \cite{vinyals2015neural,sordoni2015neural,li2016diversity,serban2016building,serban2017hierarchical,zhao2017learning,shao2017generating} present the surface string of dialogue histories by a neural encoder as the understanding results.
They suffer from severe problems, such as being uninterpretable and generating trivial responses, like ``\emph{I do not know}''.
For the first time, we propose conversational semantic roles as general-purpose representations for understanding chit-chat dialogues.
The conversational semantic roles are symbolic, alleviating the uninterpretable problem.
Comparatively, our predicates and arguments in chit-chat dialogues are reminiscent of the intent and slots for task-oriented dialogues.

A few recent efforts \cite{xing2017topic,zhang2019consistent} adopt topic modeling \cite{blei2003latent} for general dialogue understanding, where keywords are extracted from each dialogue utterance as the topic representation for that turn.
Similar to conversational SRL, topic modeling also provides symbolic and interpretable features.
Here we take one step further by modeling structures of predicates and arguments that can capture more complex information than individual words.

\begin{figure*}[ht!]
    \centering
    \includegraphics[width=\textwidth]{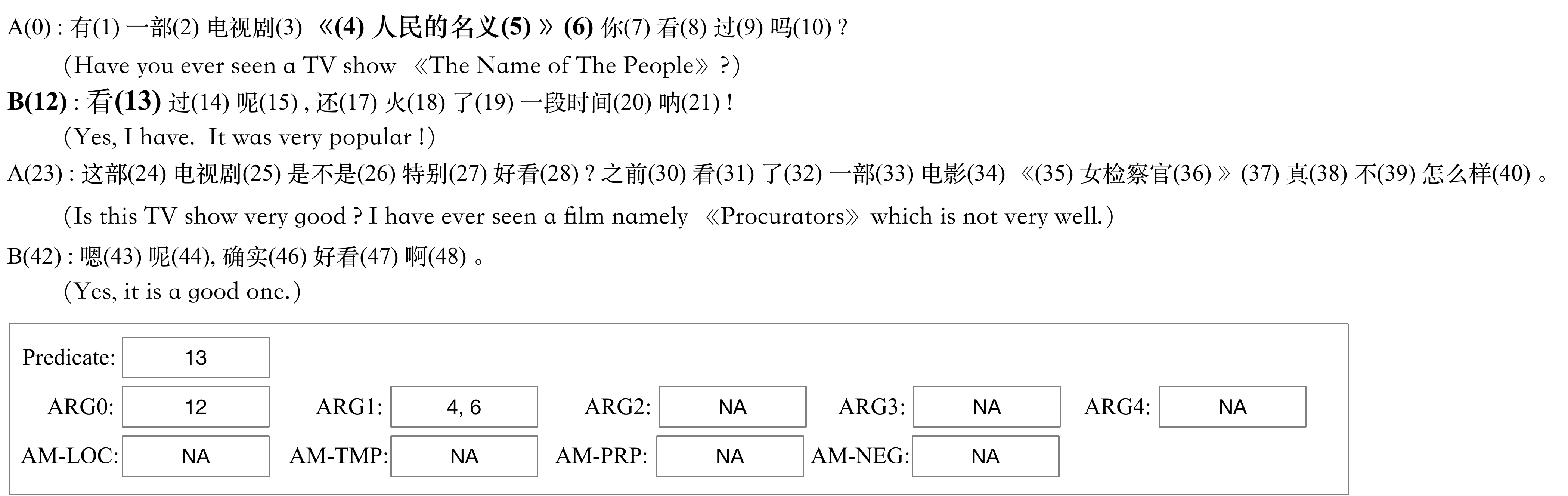}
    \caption{Interface for the annotation.}
    \label{fig:annotation_interface}
\end{figure*}

\begin{table*}[t!]
\fontsize{10}{11} \selectfont
\setlength\tabcolsep{5pt}
    \centering
    \begin{tabular}{rcccccccc}
    \toprule
        Dataset & \texttt{ARG0} & \texttt{ARG1} & \texttt{ARG2} & \texttt{ARG3} & \texttt{ARG4} & \texttt{AM-TMP} & \texttt{AM-LOC} & \texttt{AM-PRP} \\
        \midrule
        DuConv & 42.1(22.9) & 40.2(16.9) & 10.1(30.2) & 3.0(24.8) & 0.3(41.4) & 3.2(0.3) & 1.0(2.1) & 0.1(4.0) \\
        PersonalDialog & 41.4(15.1) & 40.8(8.0) & 9.6(14.5) & 1.4(22.9) & 0.1(50.0) & 5.4(1.1) & 0.9(3.2) & 0.3(0.0) \\
        NewsDialog & 42.2(26.8) & 41.7(16.3) & 7.3(20.5) & 1.4(31.1) & 0.1(83.3) & 6.1(0.2) & 0.9(0.0) & 0.3(0.0) \\
    \bottomrule
    \end{tabular}
    \caption{Percent(\%) of each type of argument and its cross-turn ratio (shown inside parenthesis).}
    \label{tab:stat}
\end{table*}

\section{Conversational SRL Dataset}
We annotate the predicate-argument structures on three datasets, including two public dialogue datasets: DuConv \cite{wu-etal-2019-proactive} and PersonalDialog \cite{zheng2019personalized}, and one in-house dataset (referred as NewsDailog), each having a \emph{very different} domain and style from the others.
Following standard SRL, we adopt the semantic roles in PropBank to label the relationships between arguments.
In this section, we first introduce the datasets (Section \ref{sec:data_duconv}) and the semantic roles based on PropBank (Section \ref{sec:data_srl}), before showing our annotation details (Section \ref{sec:data_anno}).

\subsection{Dialogue Datasets} \label{sec:data_duconv}
\paragraph{DuConv}
It is a publicly available knowledge-driven dialogue dataset, focusing on the domain of movies and stars.
It consists of 30k dialogues with 270k dialogue turns and provides a knowledge graph (KG) in the domain of movies and celebrities.
Each dialogue is collected by asking two crowd-sourced annotators to conduct a multi-turn KG grounded conversation focusing on given entities, where one plays the role of the conversation leader and the other one acts as the conversation follower.

\paragraph{PersonalDialog}
In contrast to DuConv, which involves only the domain of movies and stars, PersonalDialog includes more than 250 domains.
It is collected by crawling Weibo\footnote{A Chinese version of Twitter, \url{http://www.weibo.com}.} posts and comments.
Specifically, when a user posts a Weibo message, other users may respond with comments that may receive further comments.
This results in a tree structure rooted in the original Weibo post. An original post and one branch of its comments between the same user pairs are regarded as a dialogue session.

\paragraph{NewsDialog}
This in-house dataset is collected in a similar way as \cite{dinan2018wizard} by asking two participants to discuss each provided document.
It also follows the setting for constructing general open-domain dialogues:
two participants engage in chitchat, and during the conversation, the topic is allowed to change naturally. 
Different from \cite{dinan2018wizard}, who use Wikipedia articles, this dataset is supported by news, i.e., the participants have access to an information retrieval system that shows the news relevant to the conversation.

\subsection{Semantic Roles} \label{sec:data_srl}
We follow PropBank \cite{carreras2005introduction}, the most widely used standard for annotating predicate-argument structures. It has 32 standard semantic roles.
By analyzing the conversation dataset, we adopt 9 core semantic roles in our dialogue SRL:\\
$\bullet$ Numbered arguments (\texttt{ARG0}-\texttt{ARG4}): Arguments defining verb-specific roles. Their semantics depends on the verb and the verb usage in a sentence or verb sense. In general, \texttt{ARG0} stands for the \textit{agent} and \texttt{ARG1} corresponds to the \textit{patient} or \textit{theme} of the proposition, and these two are the most frequent roles. Numbered arguments reflect either the arguments that are required for the valency of a predicate or if not required, those that occur with high frequency in actual usage.\\
$\bullet$ Adjuncts: General arguments that any verbs may take optionally. In PropBank, there are 13 types of adjuncts, while in our dataset we only consider the most frequent four types of adjuncts, i.e., \texttt{AM-LOC}, \texttt{AM-TMP}, \texttt{AM-PRP} and \texttt{AM-NEG}. Specifically, the locative modifiers (\texttt{AM-LOC}) indicate where the action takes place. The temporal arguments (\texttt{AM-TMP}) show when an action takes place, such as {\small很快} (soon) or {\small马上} (immediately). Note that, the adverbs of frequency (e.g., {\small偶尔} (sometimes), {\small总是} (always)), adverbs of duration (e.g., {\small过两天} (in two days)) and repetition (e.g., {\small又} (again)) are also labeled as \texttt{AM-TMP}. Purpose clauses (\texttt{AM-PRP}) are used to show the motivation for an action. Clauses beginning with {\small为了} (in order to) and {\small因为} (because) are canonical purpose clauses. \texttt{AM-NEG} is used for elements such as `{\small没有}' (not) and `{\small 绝不}' (no longer).

\subsection{Annotation Details} \label{sec:data_anno}
There are two main types of semantic roles: span-based \cite{ouchi2018span,tan2018deep} and dependency-based \cite{li2019dependency}.
The former involves the start and end boundaries for each component, and the latter only considers the headword in a dependency tree for each component.
We follow the span-based form, which has been adopted by most previous works.

\begin{figure*}
    \centering
    \includegraphics[width=1.0\linewidth]{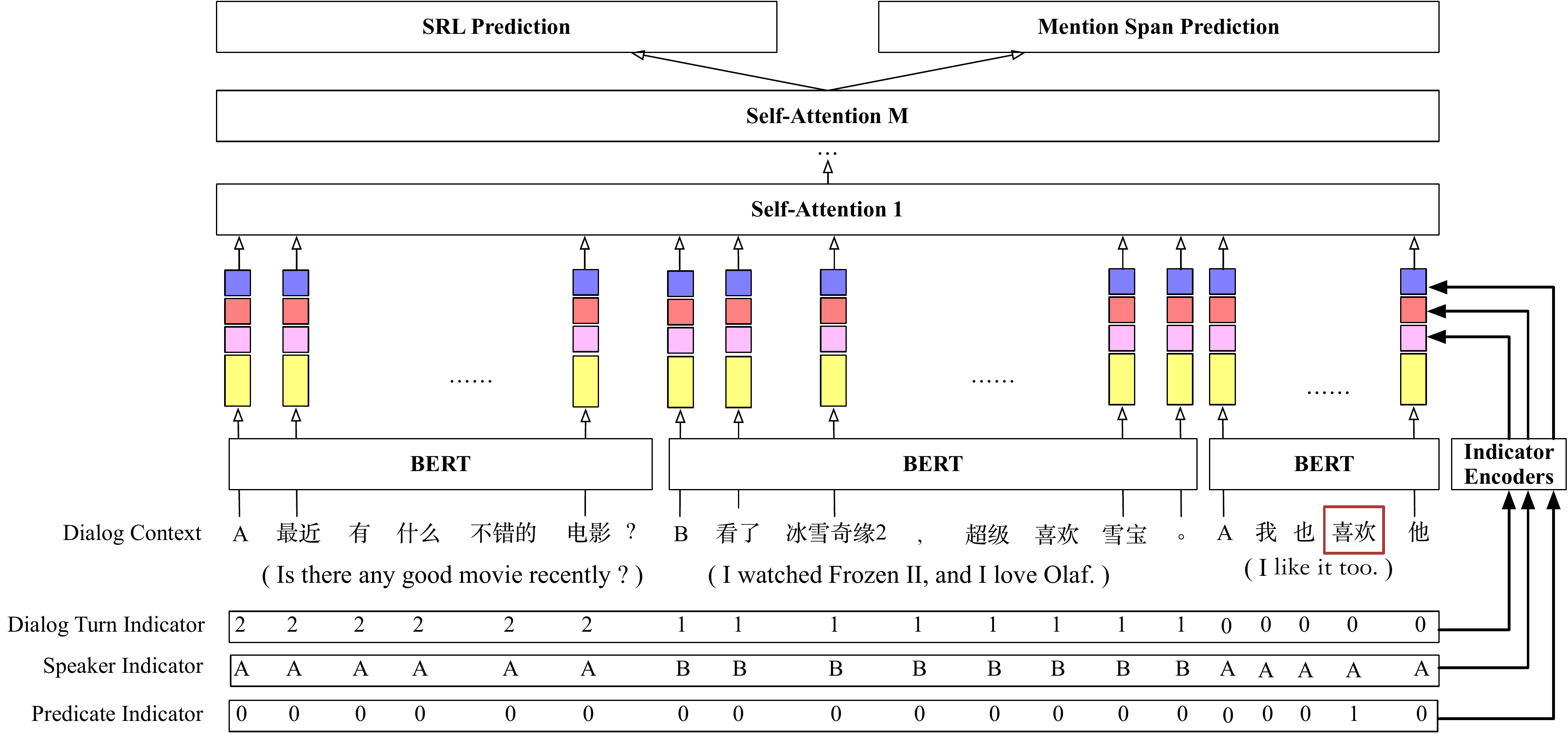}
    \caption{Model architecture. It takes the sentences from Figure \ref{fig:example} as the inputs, where the associated predicate (``喜欢(like)'') is in bold font.}
    \label{fig:model}
\end{figure*}

\paragraph{Preprocessing}
For each dialogue session, we first convert it to a paragraph by concatenating each utterance in the dialogue history.
To explicitly indicate each utterance's speakers, we insert the corresponding speaker token at the beginning of each utterance before the concatenation.
We then use Stanford CoreNLP \cite{manning-EtAl:2014:P14-5} for sentence segmentation, tokenizing, and POS-tagging.
We identify verbs by POS tagging with heuristics to filter out auxiliary verbs.

\paragraph{Labeling instructions}
We ask five annotators who are familiar with PropBank semantic roles to annotate these dialogue sessions.
%\footnote{More details about the annotation interface are shown in the Appendix.}
Following the span-based annotation standard, annotators label the index ranges for each predicate and its arguments.
In contrast to the standard sentence-level SRL, conversational SRL aims to additionally address the ellipsis and anaphora problems, which frequently occurred in the dialogue scenario.
To this end, the annotators are instructed that a \textit{valid} annotation must satisfy the following criteria: 
(1) the argument should only appear in the current or previous turns; 
(2) the argument should not be assigned to a pronoun unless its reference could not be found in previous turns;
(3) if the argument is the speaker or listener, it should be explicitly assigned to the special token we used to indicate the speaker (i.e., A or B).
(4) in cases when there are multiple choices for labeling an argument, we select the one closest to the predicate.
Figure~\ref{fig:annotation_interface} illustrates the annotation interface.
The number in the bracket indicates the token index within the dialogue paragraph.

\subsection{Statistics}
We annotated 3,000 dialogue sessions from DuConv (33,673 predicates in 27,198 utterances), 300 sessions from PersonalDialog (1,441 predicates in 1,579 utterances)
and 200 sessions from NewsDialog (3,621 predicates in 6,037 utterances).
These datasets contain 21.89\%, 12.56\%, and 20.01\% arguments that are not in the same turn with their predicates.
Also, 9.46\%, 13.05\%, and 8.27\% arguments are speaker arguments in these datasets, respectively. The average number of dialogue turns in three datasets are 9.0, 7.9 and 20.1, respectively. The average number of session tokens are 96.0, 100.1 and 230.1, respectively.

Table \ref{tab:stat} analyzes our datasets by listing the percent of each argument type and its cross-turn ratio.
For all the three datasets, arguments \texttt{ARG0}, \texttt{ARG1} and \texttt{ARG2} count for the major proportion of the arguments.
For adjunct-type arguments, \texttt{AM-TMP} and \texttt{AM-LOC} appear more than \texttt{AM-PRP}.
It is likely because humans tend to avoid mentioning reasons for simplicity.
Besides, the adjunct-type arguments have very low cross-turn ratios.
This fits our intuition that humans usually mention the time and location when describing an event or a piece of news.
For the following experiments, our annotations on DuConv are split into 80\%/10\%/10\% as train/dev/test set,
and the annotations on both PersonalDialog and NewsDialog serve only as test examples.

\section{Task Definition}
Formally, the input is a dialogue $C = (u_{1}, ..., u_{T})$ of $T$ utterances, where $u_{t} = (w_{t,1}, ..., w_{t,|u_{t}|})$ consists of a sequence of words, and $u_{T}$ is the current turn.
Following previous work \cite{tan2018deep,shi2019simple}, we cast the task into a sequential labeling problem, where a label needs to be predicted for each token in both the current turn and the dialogue history, given a predicate.
The label set draws from the cross of the standard BIO tagging scheme and the argument sets, and example labels are \texttt{B-ARG1} and \texttt{I-ARG1} that represent the beginning and intermediate token for \texttt{ARG1}, respectively.
We follow previous settings \cite{he-etal-2017-deep,tan2018deep,marcheggiani2018exploiting}, where predicate positions are already marked in both training and evaluation sets, and only focus on predicting arguments.

\section{Model}
Our model adopts a state-of-the-art system \cite{shi2019simple} for conversations.
As illustrated in Figure~\ref{fig:model}, 
it consists of an utterance encoder and $M$ self-attention layers. The input dialogue session is first fed to the utterance encoder to learn the contextual representations for each utterance.
Then a stack of self-attention layers is further used to capture the dialog-level contextual representations.
The topmost layer is the \textit{softmax} classification layer.

\paragraph{Utterances Encoder.}
Given an input dialogue $C$, we first use a pre-trained language model, namely RoBERTa \cite{liu2019roberta}, to encode each utterance\footnote{\small \url{https://github.com/brightmart/roberta_zh}.}. This results in a series of word representations, which are supposed to capture the contextual information of words within utterances.

\paragraph{Indicator Embedding.}
In addition to the utterance encoding, we also consider three types of indicator features closely related to our task.
Conceptually, the first two indicators are mainly used to explicitly encode the speaker and dialogue turn information.
Specifically, the first one is the \textbf{\textit{speaker}} indicator capturing the speaker of each word.
In our case, the speaker is either $A$ or $B$\footnote{The first speaker in a dialogue is viewed as speaker $A$.}.
The second type is the \textbf{\textit{dialog-turn}} indicator, which calculates the relative distance in terms of dialogue turn between a word to the predicate one.
The last one is the \textbf{\textit{predicate}} indicator, which is typically used in the standard SRL to distinguish the predicate words from non-predicate ones.
Here, we mark the predicate word as $1$ while other words are marked as $0$.
These indicator embeddings are first concatenated to the contextual representation from the utterance encoder and then fed to the subsequent self-attention layers.

\paragraph{Self-Attention.}
We use self-attention \cite{vaswani2017attention} to correlate the context of all dialogue histories.
In particular, we first map the matrix of input vectors  $\textbf{H} \in \mathbb{R}^{N\times d}$ ($N$ is the total number of words in $C$; $d$ represents the mapped dimension) to queries ($\textbf{Q}$), keys ($\textbf{K}$) and values ($\textbf{V}$) matrices by different linear projections:
\begin{equation}
\nonumber
    \left[
    \begin{array}{l}
            \textbf{K}\\
            \textbf{Q}\\
            \textbf{V}
    \end{array}
    \right] = 
    \left[
    \begin{array}{l}
            \textbf{W$_{k}$H}\\
            \textbf{W$_{q}$H}\\
            \textbf{W$_{v}$H}
    \end{array}
    \right]
\end{equation}
The attention weight is then computed by dot product between \textbf{Q}, \textbf{K} and the selft-attention output $\textbf{C}\in \mathbb{R}^{N\times d}$
is a weighted sum of values $\textbf{V}$:
\begin{equation}
\nonumber
\textbf{C} = {\rm Attention}(\textbf{Q}, \textbf{K}, \textbf{V}) = \texttt{Softmax}\left(\frac{\textbf{Q} \textbf{K}^{T}}{\sqrt{d_{k}}}\right)\textbf{V}
\end{equation}
where \textbf{C} = (\textbf{c$_{1}$}, ..., \textbf{c$_{N}$}) are the word representations that capture both the utterance- and dialogue-level contextual information; $d_{k}$ represents the dimension of keys.
Finally, predictions are made using a multi-layer perceptron with one hidden layer over the label set: $t_i = \mathtt{softmax}(\mathtt{MLP}(\boldsymbol{c}_i))$.

\paragraph{Training}
We train our model using two types of losses: SRL loss and mention span loss.
Both two types of losses are conditionally independent and are calculated with different linear layers, given the deep features produced by our model, and the final loss for training is their combination.

Specifically, given a set of training instances, each containing a dialogue session $C$ = \{$w_1, \dots, w_N$\}, where $N$ represents the total number of words in the dialog, we train our model with a cross-entropy loss between the gold-standard labels $y$ = \{$y_{1}, y_{2},...,y_{N}$\} and model distribution.
In addition, we use the loss for predicting mention spans to better establish the argument boundaries:
\begin{equation}
\nonumber
\begin{split}
    \mathcal{L} = - \frac{1}{N} \sum_{t=1}^{N} \underbrace{\log p(y_{t}|w_t,C;\bm{\theta})}_{\mathcal{L}_{SRL}} +  \underbrace{\log p(z_{t}|w_t,C;\bm{\theta})}_{\mathcal{L}_{span}} 
\end{split}
\end{equation}
where $\bm{\theta}$ represents the model parameters, and each $z_t$ is the gold mention-span label (also in BIO scheme) for $w_t$.
Since DuConv already provides the relevant KB fragment for dialogs, the mentions are generated by matching the KB entities with dialogue utterances. 

\section{Applications}
In this paper, we investigate whether CSRL could benefit two major conversation-based tasks, i.e., multi-turn dialogue rewriter and dialogue generation.
For each task, we first build a BERT-based baseline which already outperforms previous approaches, and then conduct experiments based on them.
In this section, we first discuss the motivation of applying CSRL into these tasks before introducing how to incorporate CSRL features into baseline models.

% problem formulation of two tasks and then introduce the method of using CSRL into two tasks.

\begin{figure*}
    \centering
    \includegraphics[width=1.0\linewidth]{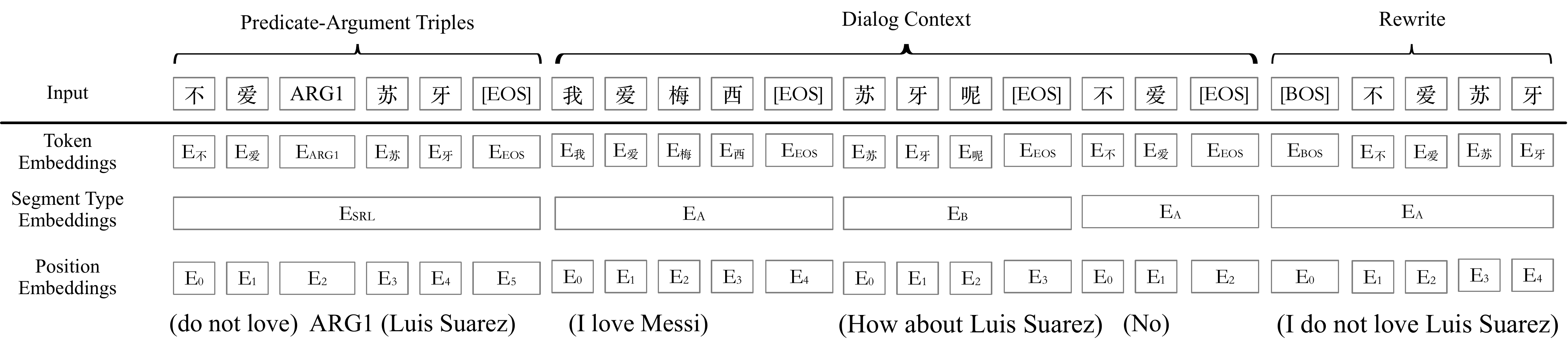}
    \caption{The input representation of the rewriter and dialogue model.}
    \label{fig:rewriter_model}
\end{figure*}
\subsection{Multi-turn Dialogue ReWriting}
Recent work on dialogue generation \cite{serban2017hierarchical,zhao2017learning,shao2017generating} has achieved impressive progress for making single-turn responses, while producing coherent multi-turn replies still remains challenging.
%One important reason is coreference and information omission, where mention is dropped or replaced by a pronoun for simplicity.
One important reason can be the long-range dependency issue caused by the lengthy context of the multi-turn scenario, where a generated reply needs to be coherent with every previous turn rather than only the last turn.
To tackle this problem, the task of sentence rewriting was introduced to alleviate the long-range issue by merging relevant historical information into the latest turn, simplifying the original problem into a single-turn problem.

Conceptually, the current state-of-the-art rewriting models follow the conventional encoder-decoder architecture that first encodes the
dialogue context into a distributional representation and then decodes it to the rewritten utterance.
Their decoders mainly use global attention methods
that attend to all words in the dialogue context without prior focus, which may result in inaccurate concentration on some dispensable words.
This observation is expected since if the text is lengthy, it would be quite difficult for deep learning models to understand as it suffers from noise and pays vague attention to the text components.

\begin{table}[t!]
\small
\centering
\begin{tabular}{l|l}
\toprule[0.8pt]
Utterance 1 & 需要粤语 \\
 & (I may need Cantonese.) \\
Utterance 2 & $\underline{\text{粤语}}$$_{\textbf{ARG0}}$是$\underline{\text{普通话}}$$_{\textbf{ARG1}}$吗 \\
& (Is Cantonese Mandarin ?) \\
Utterance 3 & $\underline{\text{不算}}$$_{\textbf{predicate}}$吧 \\
& (Maybe Not.) \\
\textbf{\textcolor{blue}{Utterance 3$^\prime$}} & \textbf{\textcolor{blue}{粤语不算普通话吧}} \\
& (Cantonese may be not Mandarin.) \\
\toprule[0.8pt]
\end{tabular}
\caption{One example of multi-turn dialogue. The goal of dialogue rewriting is to rewrite utterance 3 into 3$^\prime$.}
\label{tab:intro_examples}
\end{table}

Since CSRL could identify predicate-argument structures of a sentence, we believe that it can help pick out the important words that are semantically most related to the latest utterance that needs to be rewritten.
For example, in Table~\ref{tab:intro_examples},
our CSRL parser can find that the \texttt{\small ARG0} and \texttt{\small ARG1} of ``{\small 不算}''(is not) are ``{\small 粤语}''(Cantonese) and ``{\small 普通话}''(Mandarin), respectively.
%Consequently, our rewriting model can generate the correct output (utterance 3$^\prime$) by merging this important information from dialogue history.
%, which covers all dropped information.
Consequently, CSRL could guide the rewriter model to pay more attention to the semantically important words in the dialogue history, especially the omitted information that appears in previous turns.
Motivated by this observation, we attempt to apply CSRL into this task by first taking the CSRL parser to recognize the predicate-argument structures from dialog contexts and encode those structural information into the rewriter model. \\
\\
\textbf{Problem Formulation.}
Formally, an input for dialogue rewriting is a dialogue session $c = (u_{1}, ..., u_{N})$ of $N$ utterances, where $u_{N}$ is the most recent utterance that needs to be revised.
The output is $r$, the resulting utterance after recovering all coreference and omitted information in $u_{N}$.
Our goal is to learn a model that can automatically rewrite $u_{N}$ based on the dialogue context.\\
\\
\textbf{ReWriter Model.}
Given a dialogue context $c$, we first apply an SRL parser to identify the predicate-argument structures $z$; then conditioned on $c$ and $z$, the rewritten utterance is generated as $p(r|c, z)$.
The backbone of our infrastructure is similar to the transformer blocks in \cite{dong2019unified},
which supports both bi-directional encoding and uni-directional decoding flexibly via specific self-attention masks.
Specifically, we concatenate $z$, $c$ and $r$ as a sequence, feeding them into our model for training; during decoding, our model takes the $z$ and $c$ before generating the rewritten utterance word by word.
Our model uses a pre-trained Chinese RoBERTa \cite{liu2019roberta} for rich features.\\
\\
\textbf{Input Representation.}
For each token, its input representation is obtained by summing the embeddings for word, semantic role and position.
One example is shown in Figure~\ref{fig:rewriter_model} and details are described in the following:
\begin{itemize}
    \item The input is the concatenation of PA structures, dialog context, and rewritten utterance.
            Note that a PA structure is essentially in a tree format, where the root is a predicate and its children are corresponding semantic arguments.
            For the linearization, we decompose each PA structure into several triples
            of the form $<$\textit{predicate}, \textit{role}, \textit{argument}$>$ and concatenate them in a random order.
            A special end-of-utterance token (i.e., [EOS]) is appended to the end of each utterance for separation.
            Another begin-of-utterance token (i.e., [BOS]) is also added at the beginning of the rewritten utterance.
            The final hidden state of the last token in the final layer is used to predict the next token during generation.
    \item We expand the segment-type embeddings of BERT to distinguish different types of tokens.
            In particular, the type embedding E$_{A}$ is added for the rewritten utterance, as well as dialogue utterances generated by the same speaker in the context; the type embedding E$_B$ is used for the other speaker; E$_{SRL}$ is used as the type embedding of the tokens in predicate-argument triples.
            Position embeddings are added according to the token position in each utterance.
            The input embedding is the summation of word embedding, segment embedding, and position embedding.
\end{itemize}
\textbf{Attention Mask.}
Similar to TransferTransfo \cite{wolf2019transfertransfo}, we apply a future mask on the rewritten sequence; that is, the tokens in the rewritten utterance only attend on previous tokens in self-attention layers.
Recall that we linearize a PA structure into a concatenation sequence of triples.
Since these triples are randomly ordered, it may inevitably introduce noisy information when using a sequence encoder.
To better reflect its structural information, we elaborate the attention mask on PA sequence:
the tokens in the same PA triple have bidirectional attentions while tokens in different PA triples can not attend to each other.
And the position embeddings of tokens in the PA sequence are added according to their positions in each distinct triple rather than the total PA sequence.
In experiments, we find using these two designs helps our model to more efficiently use the SRL information.\\
\\
\textbf{Training}
We employ the NLL loss to train our model:
\begin{equation}
\nonumber
    \mathcal{L}  = - \sum_{t=1}^{T} \log p(r_t|c, z, r_{< t}; \bm{\theta})
\end{equation}
where $\bm{\theta}$ represents the model parameters, $T$ is the length of the target response $r$,
and $r_{< t}$ denotes previously generated words.

\begin{table*}[ht!]
\fontsize{10}{11} \selectfont
\setlength\tabcolsep{5pt}
\centering
\begin{tabular}{lcccccccccccc}
\toprule[0.8pt]
\multirow{2}{*}{Method} & \multicolumn{3}{c}{DuConv} & & \multicolumn{3}{c}{PersonalDialog} & & \multicolumn{3}{c}{NewsDialog} \\
\cline{2-4} \cline{6-8} \cline{10-12}
& F1$_{all}$ & F1$_{cross}$ & F1$_{intra}$ & & F1$_{all}$ & F1$_{cross}$ & F1$_{intra}$ & & F1$_{all}$ & F1$_{cross}$ & F1$_{intra}$ \\
\hline
SRL-\textit{single} & 42.8 & 0.0 & 51.2 & & 44.9 & 0.0 & 50.8 & & 54.0 & 0.0 & 61.6 \\
SRL-\textit{entire} & 42.7 & 1.1 & 51.2 & & 42.5 & 1.8 & 49.0 & & 52.5 & 1.0 & 60.2 \\
SRL-\textit{rewrite} & 39.5 & 3.8 & 49.2 & & 37.5 & 0.5 & 44.9 & & 44.0 & 2.0 & 54.6 \\
SRL-\textit{coref} & 40.9 & 8.1 & 49.4 & & 40.0 & 1.1 & 46.5 & & 45.1 & 4.2 & 54.3  \\
SRL-DuConv-\textit{entire} & 48.1 & 1.2 & 58.0 & & 40.6 & 1.2 & 42.3 & & 50.3 & 0.5 & 57.8 \\
\hline
\textbf{CSRL} &  85.7 & 78.3 & 86.9 & & 67.4 & 32.0 & 70.8 & & 74.0 & 46.2 & 78.0\\
\emph{~~~~~w/ pretrain} & \textbf{85.9} & \textbf{78.5} & \textbf{87.0} & & 
\textbf{68.7} & \textbf{34.5} & \textbf{71.9} & & \textbf{74.7} & \textbf{48.4} & \textbf{78.3} \\
\emph{~~~~~w/o span loss} & 85.4 & 78.0 & 86.6 & & 66.8 & 31.2 & 68.3 & & 72.9 & 46.0 & 76.6 \\
\toprule[0.8pt]
\end{tabular}
\caption{Evaluation results on the DuConv, PersonalDialog and NewsDialog datasets.}
\label{tab:results}
\end{table*}

\subsection{Multi-turn Dialogue Response Generation}
In addition to \emph{multi-turn dialogue rewriting}, we also conduct a preliminary study on multi-turn dialogue response generation, one of the main challenges in dialogue community.
In contrast to the single-turn dialogue response generation,
the ellipsis and anaphora may more frequently occur in multi-turn dialogues.
However, previous approaches simply concatenate each multi-turn dialogue session as a ``long'' input utterance or adopt a hierarchical neural structure without
analyzing the semantic information among these utterances.
Intuitively, CSRL could naturally extract the semantic information which could help the model to better understand the dialogue context.

In this paper, we study the effect of CSRL by simply providing it as part of the input.
Specifically, given a dialogue context $c$, we first apply our CSRL parser to identify the predicate-argument structures $z$; then conditioned on $c$ and $z$, the response is generated as $p(r|c, z)$.
The backbone of our infrastructure is similar to the transformer blocks in \cite{dong2019unified},
which supports both bi-directional encoding and uni-directional decoding flexibly via specific self-attention masks.
Specifically, we concatenate $z$, $c$ and $r$ as a sequence, feeding them into our model for training; during decoding, our model takes the $z$ and $c$ before generating the response word by word.
Our model uses a pre-trained Chinese RoBERTa \cite{liu2019roberta} for rich features.
\\
\\
\textbf{Input Representation.}
As shown in Figure~\ref{fig:rewriter_model}, the input representation of each token is obtained by summing the embeddings for word, semantic role and position.
\\
\\
\textbf{Attention Mask.}
Similar to the rewriter model, the tokens in the response only attend to previous tokens in self-attention layers; we adopt the elaborated attention mask design for the tokens in PA structures.
\\
\\
\textbf{Training.}
We employ the same NLL loss as the rewriter model to train the dialogue model.

\section{Experiments}
We conduct three types of experiments to evaluate the usefulness of CSRL.
Specifically, the first experiment evaluates the performance of the CSRL parser, while the remaining two experiments investigate the performance improvement on the dialogue rewriting and generation task that could be achieved by introducing the CSRL information.

\subsection{CSRL Evaluation}
\noindent\textbf{Baselines.}
We compare our model with baselines that explore major solutions to extend standard SRL on dialogues.
All baselines have the same neural architecture and are trained with the same loss as our model.
The only difference is that they are trained on the CoNLL-2012 benchmark \cite{pradhan2013towards}, which contains roughly 117,089 standard SRL instances. More details are listed below:
\begin{itemize} % \\ $\bullet$ 
    \item SRL-\textit{single}/-\textit{entire}: After training, both models are directly applied on our three test sets for evaluation. SRL-\textit{single} just feeds each dialogue turn to the model without considering the dialogue history, and SRL-\textit{entire} concatenates the dialogue history and current turn as a single utterance by replacing all intermediate periods with commas. % \\ $\bullet$ 
    \item SRL-\textit{rewrite}: 
    This method first employs a pre-trained conversation-based ReWriter model \cite{su-etal-2019-improving} to refine dialogue utterances so that the omitted information that occurred in previous turns could be retained in the resulting utterances. Then we use the same standard-SRL model to predict predicate-argument structures. % \\ $\bullet$ 
    \item SRL-\textit{coref}: This method first uses the same standard SRL model to predict predicate-argument structures and then applies a state-of-the-art model \cite{joshi-etal-2019-bert} for coreference resolution to find the original mention for each argument. % \\ $\bullet$ 
    \item SRL-DuConv-\textit{entire}: This model is mostly identical with SRL-\textit{entire}, except that it is trained only with the instances where every predicate-argument structure is in the same sentence. It is introduced for examining the potential annotation bias that could favor our model.
\end{itemize}

\noindent\textbf{Experimental Settings.}
We take the following settings for our models and baselines.
We train our model (CSRL) on the training portion of DuConv, containing roughly $21,600$ instances.
The dimensions of the predicate, speaker, and dialog-turn indicator embeddings are $10$, $50$, and $50$, respectively.
The hidden size $d$ of the self-attention layer is set to $1024$, and the number of heads is set to $8$.
We use four self-attention layers and tune all the hyper-parameters on the development set.

Following the previous research of standard SRL, we evaluate our system on the micro-averaged F1 over the (predicate, argument, label) tuples.
However, unlike sentence-level SRL systems, which only calculate F1 over all arguments (i.e., F1$_{all}$), we also calculate F1 over those arguments in the same, and different dialogue turns as predicates, namely F1$_{intra}$ and F1$_{cross}$. We refer these two types of arguments as \textit{intra-} and \textit{cross-}arguments, respectively.
For the SRL-\emph{rewrite} baseline, an argument is considered as a \emph{cross}-argument if it only appears in the rewritten sentence, not in the original one (before rewriting).

\begin{table*}[ht!] 
% \small
\fontsize{10}{11} \selectfont
    \centering
    \begin{tabular}{lcccccccc}
    \toprule
        Method & \texttt{ARG0} & \texttt{ARG1} & \texttt{ARG2} &\texttt{ARG3} & \texttt{ARG4} & \texttt{AM-TMP} & \texttt{AM-LOC} & \texttt{AM-PRP}  \\
    \hline
        SRL-\emph{single} & 43.4 & 49.1 & 2.6 & 0.0 & 0.0 & 74.9 & 30.4 & 50.5  \\
        SRL-\emph{entire} & 43.1 & 48.7 & 3.2 & 0.0 & 0.0 & 73.6 & 29.6 & 43.1  \\
    \hline
        CSRL \emph{w/ pretrain} & 90.1 & 85.4 & 75.4 & 72.8 & 46.4 & 82.9 & 79.2 & 34.4 \\
    \bottomrule
    \end{tabular}
    \caption{Detailed scores on the test set of DuConv dataset.}
    \label{tab:results_detail}
\end{table*}

\begin{table*}[ht!] \small
    \renewcommand{\arraystretch}{1.3}
    \centering
    \begin{tabularx}{\linewidth}{cX}
    \toprule
        {\small A}: & {\small 很快(soon)~到(is)~我们的(our)~结婚(wedding)~纪念日(anniversary)了，我们(we)~准备(prepare)~出去(go for)~旅游(travel)~！} \\
        {\small B}: & {\small 带着(with)~孩子(kid)~旅游(travel)~是不是(whether or not)~要带(bring)~很多(many)~东西(stuff)啊~！} \\
    {\small A}: & {\small 是啊(yes)，我们(we)~专门(specially)~买了(bought)~一个(a)~大箱子(suitcase)，不过(but)~\textbf{还是}(are still)~太多了(too many)，根本(at all)~放不下(cannot fit)~。} \\
        \cdashline{1-2}[0.8pt/2pt]
         \textbf{Predicate} & \textbf{Gold and Predicted Arguments}  \\
         还是 & (\texttt{ARG0}, 东西), (\texttt{ARG1}, 太多了)~~~$\vert$ $\vert$ $\vert$~~~(\texttt{ARG0}, 箱子), (\texttt{ARG1}, 太多了) \\
    \bottomrule
    \end{tabularx}
    \caption{Example dialogues and predicted argument structures by our model.}
    \label{tab:error_examples}
\end{table*}

\noindent\textbf{Main Results.}
Table~\ref{tab:results} summarizes the results of all compared models on the DuConv, PersonalDialog and NewsDialog datasets.
We can see that our model outperforms all the baselines by large margins even for the two out-of-domain datasets, where the performance gaps are 22.5 and 20 absolute F1 points, respectively.
Although the standard CoNLL-2012 benchmark has many more instances, our comparisons suggest that these corpora do not provide sufficient knowledge for the dialogue scenario, where arguments usually occur in different utterances that can be far from their predicates.
Training only with the single-sentence annotations of DuConv, SRL-DuConv-\textit{entire} is better than SRL-\textit{entire} on DuConv test set, but it is much worse on the other test sets.
The reason can be that SRL-DuConv-\textit{entire} benefits from the in-domain advantage on DuConv, but it suffers from less training data on the other two test sets.
This also indicates that the improvements of CSRL are not due to any potential annotation bias.

In addition to the standard SRL setting (SRL-\textit{single}), we also investigate other potential solutions for tackling conversations.
In contrast to SRL-\textit{single}, which only uses the single-turn information,
SRL-\textit{entire} incorporates the whole dialogue history for sequence annotation.
Intuitively, considering more context could allow the model to better find \textit{cross-}arguments.
However, we do not observe SRL-\textit{entire} achieves better performance on these datasets.
In fact, the contextual information can even hurt F1$_{intra}$ since those contextual information may introduce the noise in detecting \textit{intra-}arguments.

Introducing a state-of-the-art rewrite component, i.e., SRL-\textit{rewrite}, can slightly improve the numbers on F1$_{cross}$, but it significantly hurts F1$_{intra}$, resulting in much worse performances on F1$_{all}$.
This observation conflicts with our intuition; that is, the rewrite model could recover all co-referred and omitted mentions by rewriting the current utterance.
By analyzing $300$ rewritten DuConv utterances by the rewrite model,
we find that although the rewrite model can recover a few omitted mentions,
the grammar of rewritten utterance is usually problematic.
As a result, the standard SRL model trained on the normal corpus may fail in such noisy sentences.

Leveraging a state-of-the-art model \cite{joshi-etal-2019-bert} to generate coreference chains, SRL-\emph{coref} gets an absolute 2.4\% improvement on average over SRL-\emph{rewrite} for the \emph{cross} arguments, and SRL-\emph{coref} is also slightly better than SRL-\emph{rewrite} for the \emph{intra} cases.
However, it is still much worse than SRL-\emph{single} and SRL-\emph{entire}.
After an investigation, we find that it is mostly due to the errors of coreference resolution, i.e., some correct arguments are erroneously attached to a coreference chain.

\noindent\textbf{Discussion.}
% \paragraph{Indicator Embedding}
We also investigate the impact of indicator embeddings on the performance.
Experimental results show that without the dialog-turn encoding, the F1$_{all}$ performance of our model decreases to 82.8, 63.2, and 70.0 on DuConv, PersonalDialog, and NewsDialog, respectively.
In particular, we find that removing the dialog-turn feature severely hurts the performance on \textit{cross-}arguments.
This result is expected since the dialog-turn feature essentially reduces the distance of words in different turns, which allows the model to more easily find the arguments that occurred in previous dialogue turns.
Recall that, in the conversational SRL, the underlying speaker itself could also be a potential argument. Integrating the speaker feature is also observed to improve the performance on such arguments.

% \vspace{-0.2cm}
% \paragraph{NER Loss}
% Similar to existing standard-SRL models, one of the main challenges for our model is to establish the boundaries of entity mentions.
% By analyzing 300 random incorrect predictions of our model, we find that boundary errors contribute to 38.2\% of the total span exact match errors.
% In experiments, we observe that without the NER loss, the F1$_{all}$ performance drops at least 1.2 on all datasets.
% Although the NER loss could alleviate this problem, we believe that using a high-quality entity linking tool could further improve the performance, which we leave for the future work.

% \paragraph{Benchmark pretraining}
One may raise a natural question on whether existing standard SRL benchmarks are helpful for further improvement for our model, especially on F1$_{intra}$.
To investigate this, we first pre-train our model on the training set of CoNLL~2012 ($117,089$ examples) and fine-tune it on our annotated DuConv training set.
From Table~\ref{tab:results}, we can see that our model can benefit from the pre-training (CSRL \emph{w/ pretrain} vs CSRL), achieving better performance on all datasets.

We additionally investigate the impact of span loss to CSRL performance.
From Table~\ref{tab:results}, we can see that without the span loss, the performance of CSRL consistently drops on three datasets. 
These results are expected since minimizing the span loss could encourage the model to better detect the argument boundaries.

% \vspace{-0.2cm}
% \paragraph{Detailed Scores}
We list the detailed performance of our best model in Table~\ref{tab:results_detail}.
The results of the two most competitive baselines are also shown for comparison. 
We can see that our model significantly outperforms these baselines on all semantic roles except for \texttt{AM-PRP},
which may be due to its low cross-turn ratio.
Besides, the lack of sufficient training data may also contribute to this situation.
In particular, we find that the CoNLL~2012 contains 1,940 \texttt{AM-PRP} arguments, while our annotation only includes 75 appearances.
We also observe similar situations where SRL-based baselines do relatively well for \texttt{AM-TMP} and \texttt{AM-LOC}, and one likely reason is that they also have a meager cross-turn ratio.
Combining these observations, we may conclude that explicitly modeling conversational SRL is essential, because the ability for cross-turn reasoning cannot be acquired by training on sentence-level benchmarks.
%fully failed to predict for arguments \texttt{ARG3}-\texttt{ARG4},
%which mainly occur in different utterances in the dialogue scenario.

\begin{table*}[t!]
\fontsize{10}{11} \selectfont
\setlength\tabcolsep{5pt}
\centering
\begin{tabular}{l|c|c|c|c|c|c|c}
\toprule[0.8pt]
& BLEU-1 & BLEU-2 & BLEU-4 & ROUGE-1 & ROUGE-2 & ROUGE-L & EM \\ \hline
Trans-Gen & 78.18 & 70.31 & 51.85 & 83.1 & 67.84 & 81.98 & 24.12 \\
Trans-Pointer & 83.22 & 78.32 & 64.08 & 87.89 & 77.94 & 86.88 & 36.54 \\
Trans-Hybrid & 82.92 & 77.65 & 62.54 & 87.59 & 76.91 & 86.66 & 35.03\\
Su et al. \cite{su2019improving} & 85.41 & 81.67 & 70.00 & 89.75 & 81.84 & 88.56 & 46.33 \\
\hline
BERT & 88.21 & 85.17 & 75.64 & 90.73 & 84.35 & 89.47 & 57.36\\
BERT + CSRL &  \\
\quad w/ Bi-mask & 88.89 & 85.88 & 76.36 & 90.92 & 85.00 & 89.72 & 58.36 \\
\quad w/ Triple-mask & \textbf{89.66} & \textbf{86.78} & \textbf{77.76} & \textbf{91.82} & \textbf{85.87} & \textbf{90.52} & \textbf{60.49} \\
BERT + Partial-CSRL & 89.46 & 86.57 & 77.75 & 91.60 & 85.60 & 90.50 & 59.15\\
\hline
BERT + CSRL (upper bound) & 93.34 & 91.38 & 84.97 & 94.94 & 90.45 & 93.86 & 71.96 \\
\toprule[0.8pt]
\end{tabular}
\caption{Evaluation results on the rewriting task.}
\label{tab:rewriter_results}
\vspace{-3mm}
\end{table*}

\noindent\textbf{Error Analysis.}
We analyze the errors of our model and find that most errors are caused by incorrect boundary detection and incorrect argument selection.
In particular, the boundary detection errors mainly occurred in the entity mentions, while the lack of commonsense knowledge mainly causes argument selection errors.
Table~\ref{tab:error_examples} gives an example of argument selection error.
We can see that the correct \texttt{ARG0} of predicate ``{\small 还是}(are still)'' is ``{\small 东西}(stuff)'', while our model predicts it as ``{\small 箱子}(suitcase)''.
To correctly find the \texttt{ARG0} for this case, one should be aware of 
a suitcase is a \textbf{container}, and it has a capacity which can only carry limited \textbf{stuff}.
We leave incorporating such commonsense knowledge for future work.

\subsection{Multi-turn Dialogue ReWriter Evaluation}
We evaluate our enhanced ReWriter model on a rewrite dataset built by \cite{su2019improving}.
This dataset is generated by crawling multi-turn conversational data from several popular Chinese social media platforms.
Specifically, this dataset contains 17,890 examples,     which are further split as 80\%/10\%/10\% for training/development/testing, respectively.
The hyper-parameters used in our model are listed as follows.
The network parameters of our model are initialized using RoBERTa.
The batch size is set to 32.
We use Adam \cite{kingma2014adam} with learning rate 5e-5 to update parameters.
We directly employ the CSRL parser that pre-trained on the DuConv to conduct the experiments on the rewriting task.

\noindent\textbf{Results.}~~
Following previous works, we used BLEU, ROUGE, and the exact match score (EM) (the percentage of decoded sequences that exactly match the human references).
We implemented three baselines that use the same transformer-based encoder but differ in the choice of the decoder.
Specifically, \textit{Trans-Gen} uses a pure generation decoder which generates words from a fixed vocabulary;
\textit{Trans-Pointer} applies a pure pointer-based decoder \cite{vinyals2015pointer} which can only copy the word from the input;
\textit{Trans-Hybrid} uses a hybrid pointer+generation decoder as in \cite{see2017get}, which can either copy the words from the input or generate words from a fixed vocabulary.
Table~\ref{tab:rewriter_results} summarizes the results of our model and these baselines.

\noindent\textbf{Discussion.}
We can see that even without the SRL information, our model still significantly outperforms these baselines,
indicating that adopting a pre-trained language model could greatly improve the performance of such a generation task.
We can also see that the model with the pointer-based decoder achieves better performance than the generation-based and the hybrid one, which is similar to the observation as in \cite{su2019improving}.
This result is expected since there is a high chance the coreference or omission could be perfectly resolved by only using previous dialogue turns.
In addition, we find that incorporating the SRL information can further improve the performance by at 1.45 BLEU-1 and 1.6 BLEU-2 points, achieving state-of-the-art performances.

Let us first look at the impact of attention mask design on our model.
To incorporate the CSRL information into our model, we view the linearized predicate-argument structures as a regular utterance (say $u_{pa}$) and append it in the front of the input.
We experimented with two choices of attention masks.
Specifically, the first one is a bidirectional mask (referred as Bi-mask), that is, words in $u_{pa}$ could attend each other;
the second one (referred as Triple-mask) only allows words to attend its neighbors in the same triple, i.e.,
words in different triples are not visible to each other.
From Table~\ref{tab:rewriter_results}, we can see that the latter one is significantly better than the first one.
We think the main reason is that the second design independently encodes each predicate-argument triple,
which prevents the unnecessary triple-internal attentions, better mimicking  the SRL structures.

Since our framework works in a pipeline fashion, one bottleneck of our system can lie in the performance of the CSRL parser.
One natural question is how accurate our CSRL parser can be and how much performance improvement for the rewriter model we could have by introducing the CSRL information.
To investigate this, we employ a standard SRL parser\footnote{This parser is trained on the CoNLL-2012 dataset.} to analyze the gold rewritten utterance.
These extracted PA structures are \emph{considered} as the upper bound to measure the accuracy of our CSRL parser.
In particular, we evaluate our CSRL parser on the micro-averaged F1 over the (\textit{predicate}, \textit{argument}, \textit{label}) tuples.
We find our CSRL parser achieves 75.66 precision, 74.47 recall, and 75.06 F$_{1}$.
On the other hand, we use the upper bound SRL results instead of our CSRL parsing results to train and test the model (referred as BERT+CSRL (upper bound)).
From Table~\ref{tab:rewriter_results}, we can see that all evaluation scores are significantly improved.
This result indicates that the performance of our rewriter model is highly relevant to the CSRL parser,
and the performance of our current CSRL parser is still far from satisfactory, which we leave for future work.

We also investigate which type of dialogues our model could benefit from incorporating CSRL information?
By analyzing the dialogues and our predicted rewritten utterances, we find that the CSRL information mainly improves the performance on the dialogues that require information completion.
One omitted information is considered as properly completed if the rewritten utterance recovers the omitted words.
We find the CSRL parser naturally offers important guidance into the selection of omitted words.
% Examples of rewritten utterances are shown in the Appendix.

Recall that, there is one additional scope option to apply the CSRL parser to extract PA structures,
i.e., only working on the last utterance that needs to be rewritten.
We evaluate this option on our dataset (referred as BERT+Partial-CSRL) and the results are shown in Table~\ref{tab:rewriter_results}.
We can see that reducing the CSRL scope may slightly hurt the performance,
which we think is because larger CSRL scope could provide additional guidance for the rewriter model.

\begin{table}[t!] 
\fontsize{10}{11} \selectfont
    \centering
    \begin{tabular}{lccc}
    \toprule
        Method & BLEU 1/2 & DISTINCT 1/2 & Human \\
        \hline
        Wu et al. \cite{wu-etal-2019-proactive} & 0.347/0.198 & 0.057/0.155 & 2.62 \\
        Ours & \textbf{0.421/0.286} & \textbf{0.113/0.312} & \textbf{3.61} \\
        \quad w/o CSRL & 0.395/0.268 & 0.103/0.208 & 3.27 \\
    \bottomrule
    \end{tabular}
    \caption{Evaluation results of dialogue response generation models on DuConv.}
    \label{tab:results_dialog}
\end{table}

\subsection{Multi-turn Dialogue Generation Evaluation}
We evaluate our dialogue response generation model on the DuConv dataset, which includes 19,858/2,000/5,000 dialogue sessions for training/developing/testing.
Recall that, DuConv requires the dialogue model to generate the response using the given knowledge graph. We simply linearize the given knowledge graph and append it to the dialogue history, which is viewed as the dialogue context.
To evaluate the performance of the dialogue model, we leverage several common metrics, including BLEU and DISTINCT1/2.
Table~\ref{tab:results_dialog} lists the results of our model and previous approaches on DuConv.

We can see that even without the CSRL information, our BERT-based baseline could already
outperform the previous approach in terms of BLEU scores.
After introducing the predicate-argument information, the performance could be significantly improved.
In addition to the automatic evaluation criteria, we also perform human evaluation on
the generation results.
Specifically, we randomly select 400 generation results and employ three annotators
to evaluate the coherence of the response against the dialogue context by giving a score range from 1 (worst) to 5 (best). Experimental results show that in around 75\% of cases,
CSRL could improve the coherence of the generated results.
In experiments, we also find the attention mask design is critical to achieving such improvement; that is, the triple-mask is significantly better than the bidirectional mask.

\section{Conclusions and Future Work}
Standard SRL models trained on benchmarks of sentence-level annotations may fail to analyze multi-turn dialogues where ellipsis and anaphora frequently occur.
We introduced the task of conversational SRL (CSRL), where arguments can be dialogue participants or spans in the dialogue context.
To facilitate training and evaluation of data-driven methods, we collected three high-quality, manually annotated datasets.
We also designed a BERT-based model to compare CSRL with standard SRL on parsing out-of-domain dialogues.
Experimental results show that our model trained by CSRL not only significantly outperforms standard-SRL-based baselines but also achieves satisfactory robustness on domain transfer.
Further explorations of CSRL on dialogue response generation and dialogue context rewriting also demonstrate superior performances than previous state-of-the-art methods.

In the future, we will explore more applications of the CSRL in the dialogue community
and improve the robustness of the CSRL on various domains.

% or


\appendices

% \section*{Acknowledgment}

\ifCLASSOPTIONcaptionsoff
  \newpage
\fi



\bibliographystyle{IEEEtran}
\bibliography{all}


% \begin{IEEEbiography}{Kun Xu}
% \end{IEEEbiography}

% \begin{IEEEbiographynophoto}{Linfeng Song}

% % insert where needed to balance the two columns on the last page with
% %\newpage

% \begin{IEEEbiographynophoto}{Dong Yu}


%\vfill


\end{CJK*}
\end{document}